\documentclass{article}

\usepackage[preprint]{neurips_2025}

\usepackage[utf8]{inputenc} % allow utf-8 input
\usepackage[T1]{fontenc}    % use 8-bit T1 fonts
\usepackage{hyperref}       % hyperlinks
\usepackage{url}            % simple URL typesetting
\usepackage{booktabs}       % professional-quality tables
\usepackage{amsfonts}       % blackboard math symbols
\usepackage{nicefrac}       % compact symbols for 1/2, etc.
\usepackage{microtype}      % microtypography
\usepackage{xcolor}         % colors

\workshoptitle{Efficient Reasoning}

\usepackage{amsmath}
\usepackage{amsthm}
\usepackage{amssymb}
\usepackage[english]{babel}
\newtheorem{theorem}{Theorem}
\newtheorem{lemma}[theorem]{Lemma}

\usepackage{pbox}
\usepackage{adjustbox}
\usepackage{array}
\renewcommand{\arraystretch}{1.4} % extra spacing between rows

\usepackage{booktabs} % For professional-looking table rules (\toprule, \midrule, \bottomrule)

\usepackage[dvipsnames]{xcolor}
\usepackage{float}

\definecolor{rowgray}{gray}{0.96}
\definecolor{labelcolor}{rgb}{0.4, 0.2, 0}  % brownish

\usepackage{caption}
\usepackage{tcolorbox}
\usepackage[normalem]{ulem} % For the strikethrough command \sout
\usepackage{wrapfig}
\usepackage{multirow} % For spanning rows vertically
\usepackage{lipsum} % For spanning rows vertically

\usepackage{tabularx}                       % For tables with auto-adjusting column widths
\usepackage{array}

\definecolor{myblue}{RGB}{60, 120, 220}
\definecolor{myred}{RGB}{220, 40, 40}
\definecolor{boxgray}{gray}{0.95}
\definecolor{boxgray}{gray}{0.95}
\definecolor{codegray}{gray}{0.88}

\definecolor{rejectblue}{HTML}{1F77B4}
\definecolor{pivotred}{HTML}{D62728}
\definecolor{boxbg}{HTML}{F7F7F7}

\newcolumntype{L}{>{\raggedright\arraybackslash}p{0.95\textwidth}}

\newtheorem{definition}{Definition}
\newcounter{prompt}
\usepackage{tcolorbox}
\tcbuselibrary{skins,breakable,listings} % REQUIRED for tcblisting/newtcblisting

\usepackage{hyperref} % for \autoref
\newtcblisting[auto counter,number within=section]{fullwidthprompt}{
  enhanced, breakable, float*=t, width=\textwidth,
  colback=black!2, colframe=black!75, boxrule=0.4pt, arc=2pt,
  left=6pt,right=6pt,top=6pt,bottom=6pt,
  title={\normalsize Prompt~\thetcbcounter: Reasoning Soundness Check},
  fonttitle=\bfseries,
  listing engine=listings,
  listing only,
  listing options={
    basicstyle=\ttfamily\small,
    breaklines=true,
    columns=fullflexible,
    showstringspaces=false,
    upquote=true
  }
}

\usepackage[ruled,vlined,linesnumbered]{algorithm2e}

\title{Reject Only \emph{Critical} Tokens: \\
Pivot-Aware Speculative Decoding }

\author{%
Amir Ziashahabi $^{1*}$ \quad Yavuz Faruk Bakman$^{1*}$ \quad Duygu Nur Yaldiz$^1$ \\ \textbf{Mostafa El-Khamy}$^2$ \quad
\textbf{Sai Praneeth Karimireddy}$^1$ \quad \textbf{Salman Avestimehr}$^1$\\
$^1$University of Southern California \\ $^2$ Device Solutions Research America, Samsung Semiconductor, Inc.\\
\texttt{\{ziashaha,
ybakman,yaldiz,karimire,avestime\}@usc.edu}\\
\texttt{mostafa.e@samsung.com}
}

\begin{document}
\begingroup
\renewcommand\thefootnote{\fnsymbol{footnote}}
\footnotetext[1]{Equal contribution.}
\endgroup

\maketitle

\begin{abstract}
Speculative Decoding (SD) ensures that the output matches the target model's distribution exactly. 
However, we argue that this distribution matching requirement is too stringent and results in unnecessarily low acceptance rates, limiting potential speedups. 
Instead, we advocate a \textit{reformulation} of the decoding objective: the proposed decoding strategy should match the \textit{expected utility}, i.e., the task-specific performance, of the target model. 
This perspective also aligns better with real-world use cases of LLMs, where utility (e.g., code correctness, factual accuracy) is often more important than sampling distribution. 
Based on this reformulation, we propose a novel decoding strategy: \textbf{Pivot-Aware Speculative Decoding}, which rejects only those tokens that would lead to a utility drop in the final output. 
We refer to these critical tokens as \textit{pivot tokens}. We propose a method for labeling tokens as pivotal or non-pivotal and train a lightweight classifier to detect them. This method can be viewed as a relaxed version of standard SD, which offers much higher acceptance while preserving utility. 
% under reasonable assumptions. 
We evaluate our method across various datasets, demonstrating that we can achieve up to $2.5 \times$ speedup with comparable utility. Source code is available at \url{https://github.com/amir-zsh/PAD}.
\end{abstract}

\section{Introduction}
% Language models are widely used across a broad range of applications, from scientific research to chatbots \cite{openai2023gpt4}. Their strong performance is largely attributed to their massive parameter numbers and extensive training data~\cite{kaplan2020scalinglawsneurallanguage}. Consequently, substantial research and engineering efforts have been devoted to building Large Language Models (LLMs) with hundreds of billions of parameters~\cite{llama3modelcard}. 
While LLMs demonstrate impressive performance in many domains \cite{openai2023gpt4}, they also come with a major drawback: slow text generation. LLMs typically follow an auto-regressive structure, generating one token at a time in a sequential manner. This sequential nature significantly slows down generation, especially as model size increases. \textit{Speculative Decoding (SD)}~\cite{spec_decoding, chen2023accelerating} addresses this bottleneck by leveraging a small, fast \textit{draft model} alongside the large \textit{target model}. Specifically, the draft model first generates a sequence of $N$ tokens. Then, the target model verifies these $N$ tokens in parallel by computing the acceptance probability as the ratio of the sampling probabilities of $N$ tokens assigned by the target and draft models. This parallel verification allows the target model to skip sequential generation for accepted tokens, which leads to substantial speedups, which is roughly proportional to the fraction of tokens accepted. Importantly, acceptance based on sampling probability ratios allows SD to preserve the original sampling distribution of the target model, meaning that SD's output matches that of the target model.

Although SD guarantees an equivalent sampling distribution to the target model, its speedup is often limited by the low acceptance rate of draft tokens imposed by this strict distribution-matching requirement. Recent works attempt to mitigate this low-acceptance rate limitation through various strategies, such as aligning the draft model's sampling distribution more closely with the target model \cite{zhou2024distillspec}, or employing heuristics like rejecting only hallucinated, low-quality tokens \cite{bachmann2025judge}. 

\begin{wrapfigure}{r}{0.5\textwidth}
    % This adjusts the vertical spacing. You may need to change or remove it.
    % \vspace{-20pt} 
    \centering
    \begin{tcolorbox}[
        colback=boxgray,
        colframe=black,
        width=\linewidth, % Use \linewidth to fit inside the wrapfigure
        arc=2mm, 
        auto outer arc,
        % boxrule=1pt,
        % title=\textbf{Motivation: Unnecessary Rejections in Speculative Decoding},
        halign title=center,
        fontupper=\scriptsize % Make font smaller to fit better
    ]
    % --- The Question ---
    \textbf{Question:} An agricultural cooperative must ship 6500 kg of potatoes. During transport by truck, 150 kg are damaged and therefore cannot be sold. The potatoes are distributed in 50 kg bags, each bag being sold for \$72. What will the sale of the potatoes bring?
    \vspace{2mm}
    \hrule
    \vspace{2mm}

% --- The Model's Step-by-Step Answer ---
\textbf{Model Output:}

To find\ \textcolor{myblue}{ the}\ \textcolor{myblue}{ sale}\ \textcolor{myblue}{ of}\ the potatoes\textcolor{myblue}{:}

1.\ \textcolor{myblue}{Calculate} the total weight of potatoes that can be sold:

\ \ \$ 6500 - 150 = 6350  \$ kg\textcolor{myblue}{.}

2.\ \textcolor{myblue}{Calculate}\ \textcolor{myblue}{ the} number of\ \textcolor{myblue}{ bags}:

\ \ \$ 6350 $\div$ 50 = 127  \$ bags.

3.\ Calculate the total\ \textcolor{myblue}{ revenue}\ \textcolor{myblue}{ from}\ \textcolor{myblue}{ selling}\ \textcolor{myblue}{ the}\ \textcolor{myblue}{ potatoes}:

\ \ \$ 127 $\times$ 72 = 9\sout{\textcolor{myblue}{2}}\textcolor{myred}{1}44  \$.
\hspace{1cm}

\vspace{2mm}
\textbf{Final Answer:}
\texttt{9144}
\end{tcolorbox}
\caption{To match the target model's distribution, SD rejects many tokens that a draft model proposes (shown in \textcolor{myblue}{blue}). Most of these rejections are unnecessary, and fixing a single token (\sout{\textcolor{myblue}{2}} $\rightarrow$ \textcolor{myred}{1}) is enough to recover the correct answer.}
\label{fig:sd_motivation}
\end{wrapfigure}

%We argue that the strict requirement of matching the full sampling distribution of the target model may be unnecessarily restrictive for achieving high acceptance rates, and thus better speedup.

In contrast to existing approaches, we advocate \textit{reformulating SD} to prioritize what matters most in the practical use of LLMs: their downstream performance, or in other words, their \textit{utility}.  In most real-world use cases of LLMs, the exact probability distribution is often irrelevant compared to ensuring that the model's outputs serve the end task effectively, which is to achieve high utility.
% The utility of the  model can be naturally conceptualized in many verifiable tasks where evaluation is straightforward, for example, generating operational, bug-free, low complexity code for a programming task, providing factual answers in question answering, or producing accurate reasoning and results for mathematical problems. 
Motivated by this insight, we modify the objective of SD such that, instead of requiring the proposed decoding to match the target model’s sampling distribution, we aim for the proposed decoding to match the \textbf{expected utility} of the target model’s outputs. 

Following this reformulation, we propose \textbf{Pivot-Aware Speculative Decoding}, a decoding strategy that \textit{rejects only the tokens whose generation would lead to a utility drop in the final output of the target model.} We name these critical tokens as \textbf{pivot} tokens as shown in Figure \ref{fig:sd_motivation} and train a lightweight classifier to identify them effectively. This approach enables significant speedups without sacrificing utility, arguably the most important metric in many real-world tasks. Since utility is definable across a wide range of tasks, our method results in a generic and extensible decoding algorithm.

\section{Preliminaries}

A language model defines a distribution $p(\mathbf{x})$ over token sequences $\mathbf{x}=(x_1,\ldots,x_k)\in \mathcal{V}^k$, factorized autoregressively:
\[
p(\mathbf{x})=\prod_{t=1}^{k} p(x_t \mid \mathbf{x}_{<t}),
\]
where $\mathbf{x}_{<t}$ is the prefix up to $t-1$. Given a context $\mathbf{x}_c$, the model generates an output sequence $\mathbf{y}\sim p(\cdot\mid \mathbf{x}_c)$ token by token, which is slow for large models.

\emph{Speculative Decoding (SD)} accelerates sampling from a target model $p_{\text{target}}$ using a smaller draft model $p_{\text{draft}}$. At position $t$, the draft proposes a block of $\gamma$ tokens,
\[
\mathbf{x}_{t:t+\gamma-1} \sim p_{\text{draft}}(\cdot \mid \mathbf{x}_{<t}),
\]
which are verified in parallel by $p_{\text{target}}$. Each proposed token $x_{t+i}$, $0\le i<\gamma$, is accepted with probability
\[
\min\!\left(1,\; \frac{p_{\text{target}}(x_{t+i}\mid \mathbf{x}_{<t+i})}{p_{\text{draft}}(x_{t+i}\mid \mathbf{x}_{<t+i})}\right).
\]
If all $\gamma$ tokens are accepted, an additional token $x_{t+\gamma}$ is sampled from $p_{\text{target}}$. If some $x_{t+i}$ is rejected, a replacement is drawn from
\[
p'(x\mid \mathbf{x}_{<t+i})=\operatorname{norm}\!\left(\max\!\left(0,\; p_{\text{target}}-p_{\text{draft}}\right)\right),
\]
ensuring the overall sampling distribution equals $p_{\text{target}}$~\citep{spec_decoding,chen2023accelerating}. 

% Let $\tau\in[0,\gamma]$ be the average acceptance length per draft block (expected number of consecutively accepted draft tokens before a rejection or block end), and define the draft acceptance ratio
% $
% \eta \;:=\; \frac{\tau}{\gamma}\in[0,1].
% $
% Let $t_{\text{draft}}$ and $t_{\text{target}}$ denote the wall-clock time for a single forward pass of the draft and target models, respectively. The expected speedup from speculative decoding can be approximated using 
% \[
% \text{speedup} = 
% \frac{\big(\eta\gamma + 1\big)\, t_{\text{target}}}{\,\gamma\, t_{\text{draft}} \;+\; t_{\text{target}}\,}.
% \]
% The numerator is the baseline time for the target model to generate the same expected token count per verification round ($\eta\gamma+1$ tokens, including the bonus token when the block fully accepts), while the denominator is the SD time for one round (one draft pass over $\gamma$ tokens plus one target verification pass). This approximation assumes parallel verification, negligible orchestration overhead, and similar per-token cost scaling within each pass.

Let $\tau\in[0,\gamma]$ be the \emph{average acceptance length} per draft block (expected number of consecutively accepted draft tokens before a rejection or block end), and define the draft acceptance ratio
\[
\eta \;:=\; \frac{\tau}{\gamma}\in[0,1].
\]

Let $t_{\text{draft}}$ and $t_{\text{target}}$ denote the wall-clock time for one forward pass of the draft and target models, respectively. One SD block costs
\[
T_{\text{SD}} \approx \underbrace{\gamma\,t_{\text{draft}}}_{\text{draft proposes }\gamma}
\;+\;
\underbrace{t_{\text{target}}}_{\text{one parallel verify (+1) pass}}.
\]
In expectation, that block yields \emph{$\eta\gamma+1$ target tokens} (the $\eta\gamma$ accepted proposals plus the extra $+1$ when the block fully accepts). By contrast, sampling from the target alone would take $(\eta\gamma+1)\,t_{\text{target}}$ time to produce the same number of tokens. Hence the expected speedup is
\[
\text{speedup} \;\approx\;
\frac{(\eta\gamma+1)\,t_{\text{target}}}{\gamma\,t_{\text{draft}} + t_{\text{target}}}
\]
Because speedup scales with $(\eta\gamma+1)$ in the numerator, any method that increases $\eta$ directly yields more accepted tokens per verify pass and higher generation speed.

\section{Methodology: PAD (Pivot-Aware Speculative Decoding)}
\label{sec:method}
\begin{figure*}[t!]
    \centering
    \includegraphics[width=\textwidth,height=0.5\textheight,keepaspectratio]{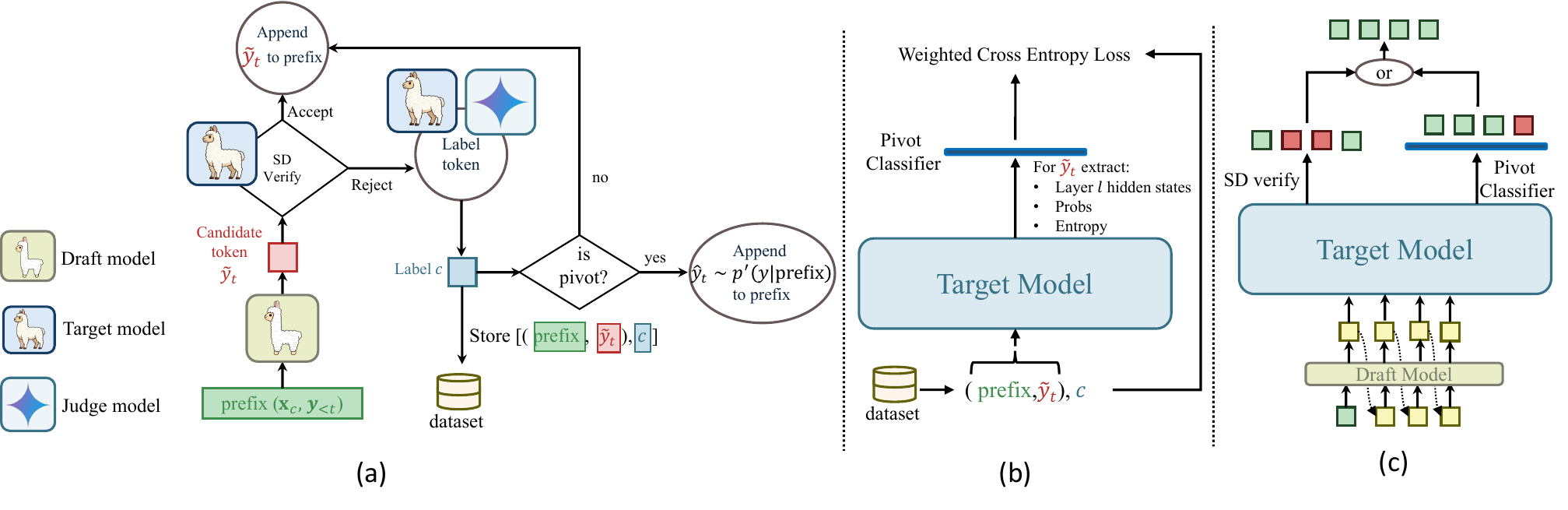}
    \caption{Pivot-Aware Speculative Decoding (PAD). (a) Dataset generation: label SD-rejected draft tokens via target-model rollouts with an LLM-as-judge sanity check; (b) Training: fit a pivot classifier on target-side features (layer-$\ell$ hidden states, logits, entropy); (c) Inference: accept tokens if standard SD accepts them, or if the classifier predicts \emph{non-pivot}.}
    \label{fig:pad}
\end{figure*}

\subsection{Objective: Matching Target Utility.}  
To improve draft token acceptance and generation speed, we propose matching the \textit{utility} of the target model rather than its sampling distribution. Utility, defined as task performance given a context or query, can be continuous, but we adopt a binary definition for simplicity.

\begin{definition}[Utility]
Let $x$ denote a query or context, and $y$ be the language model's output. Given an evaluation function $\text{Eval}(y, x)$ and a user-specified threshold $\theta_{\text{eval}}$, the utility function $u(y, x)$ is defined as:
$$
u(y, x) =
\begin{cases}
1, & \text{if } \text{Eval}(y, x) \geq \theta_{\text{eval}}, \\
0, & \text{otherwise}.
\end{cases}
$$
i.e. $u(y, x) = 1$ if the model's output achieves the desired evaluation score; otherwise, $u(y, x) = 0$. 
\end{definition}

This binary formulation naturally aligns with many tasks where correctness is clearly defined, such as programming, math, and factual question answering.

Given the definition of utility, we now aim to ensure that the proposed decoding strategy, denoted by $\hat p$, achieves an \textit{expected utility} which closely matches that of the target model $p_{\text{target}}$.
Let $U(p,x) = \mathbb{E}_{\mathbf{y} \sim  p(\cdot \mid \mathbf{x})} \left[u(\mathbf{y}, \mathbf{x})\right]$ denote the expected utility of $p$ for context $x$.

\begin{definition}[$\epsilon$-Utility preserving decoding] \label{def:utility-preserving-decoding}
Suppose we are given a task or set of tasks, with inputs $\mathbf{x}_c$ drawn from a dataset distribution $\mathcal{D}$. Then we say that $\hat p$ is $\epsilon$-utility preserving if:
\begin{equation}\label{objective}
    \mathbb{E}_{\mathbf{x}_c \sim \mathcal{D}} \left[
    U(\hat p, x_c)
 \right]
\geq 
\mathbb{E}_{\mathbf{x}_c \sim \mathcal{D}} \left[
U(p_{\text{target}},x_c )
\right] - \epsilon
\end{equation}
where $\epsilon \geq 0$ is a user-specified small tolerance value. 
%that bounds the allowable utility gap between the proposed decoding strategy and the target model. 
\end{definition}
It is trivial to verify that $p_{\text{target}}$ is utility preserving for any $\epsilon$. For this reason, utility preservation is a \textit{relaxation} of the original SD objective (which enforces exact distributional equivalence). In the next section, we describe how our proposed decoding strategy can take advantage of this relaxation in a non-trivial manner and increase the acceptance rate.

\subsection{Pivot-Aware Speculative Decoding.} Our goal is to design a new decoding strategy $p_{\text{PAD}}$ that is utility preserving as in Definition~\ref{def:utility-preserving-decoding}. To this end, we propose an approach based on rejecting only those tokens that \textit{would lead to a utility drop in the final output of the target model, assuming the remainder of the generation is completed by the target model itself}. We refer to such ``utility-changer'' tokens as \textbf{pivot tokens}. Formally, we define a pivot token as follows:

\begin{definition}[Pivot Token]\label{def:pivot-token}
Suppose we are given a context $\mathbf{x}_c$, $\mathbf{y}_{<t}$ is the tokens generated before position $t$, and $\tilde y_t$ is a candidate token. $\tilde y_{t}$ is a pivot token at time $t$ if 
\begin{equation}
    U(p_{\text{target}},(\mathbf{x}_c,\mathbf{y}_{<t}, \tilde y_{t})) 
    \leq
    U(p_{\text{target}},(\mathbf{x}_c,\mathbf{y}_{<t})) - \epsilon
\end{equation}
\end{definition}
That is, $\tilde y_t$ is a pivot token if conditioning on the $t$th token being $\tilde y_t$ results in a loss of expected utility when sampling completions (rollouts) from the target distribution. In other words, pivot tokens pivot the generation trajectory toward lower utility outputs.
Note that whether a particular token is a pivot token depends on the query $\mathbf{x}_c$, the output so far $\mathbf{y}_{<t}$, the target distribution $p_{\text{target}}$, and also the utility function $u$.

Given a binary classifier $f_{\text{pivot}}(y_t,\mathbf{y}_{<t},\mathbf{x}_c)$ indicating whether $\tilde y_t$ is a pivot token, our decoding $p_{\text{PAD}}$ follows SD with a modified rule: if $f_{\text{pivot}}(\tilde y_t,\mathbf{y}_{<t},\mathbf{x}) < \sigma$ directly accept $\tilde y_t$ otherwise follow standard SD. This modification relaxes SD's acceptance criteria by always accepting tokens deemed not pivotal ($f_{\text{pivot}}(\tilde y_t,\mathbf{y}_{<t},\mathbf{x}) < \sigma$), leading to a higher token acceptance rate.

\begin{lemma}[Rejecting only pivot tokens preserves utility]
\label{lem:recall}
If $f_{\text{pivot}}$ has $100\%$ recall on pivot tokens (i.e., it never labels a pivot as non-pivot), then $p_{\text{PAD}}$ satisfies Definition \ref{def:utility-preserving-decoding} with $\epsilon=0$.
\end{lemma}

The formal proof is provided in Appendix~\ref{sec_proof}. Note that a trivial 100\% recall pivot-token classifier can be obtained by labeling all draft tokens as pivots, which reduces to standard SD with many rejections. Classifier quality dictates the trade-off between utility preservation and speedup: better classifiers reject fewer non-pivots, increasing efficiency while maintaining utility.

\subsection{Pivot Classifier: Data and Training}
\label{sec:classifier}

Direct labeling via Definition~\ref{def:pivot-token} is intractable because, for each candidate token, it requires taking an expectation over all downstream continuations. We therefore use a Monte Carlo rollout approximation and add two safeguards to control variance and common failure modes.

\paragraph{Candidate harvesting.}
We only attempt to label tokens that \emph{standard SD would reject}. Concretely, for each $\mathbf{x}\!\sim\!\mathcal{D}$ and step $t$, draw a draft token $\tilde y_t\!\sim\! p_{\text{draft}}(\cdot\mid \mathbf{x},\mathbf{y}_{<t})$. If SD would accept $\tilde y_t$ under its usual verification rule, we skip it. Otherwise we mark it as a \emph{candidate} and proceed to label. This focuses the limited labeling budget on the ambiguous frontier where acceptance decisions actually change behavior and utility.

% \paragraph{Rollout estimate with tolerance (how to label).}
% Using the utility functional $U(p,x)$ from Section~\ref{sec:3.1}, instantiate
% \[
% p_0 \;=\; p_{\text{target}}(\cdot \mid \mathbf{x},\mathbf{y}_{<t}),
% \qquad
% p_1 \;=\; p_{\text{target}}(\cdot \mid \mathbf{x},\mathbf{y}_{<t},\tilde y_t).
% \]
% We estimate the two expected utilities via $N$ independent rollouts:
% \[
% \widehat U_0 \;\approx\; U(p_0,\mathbf{x}),
% \qquad
% \widehat U_1 \;\approx\; U(p_1,\mathbf{x}),
% \]
% where $u$ is the (binary) utility from Section~\ref{sec:3.1}. Because both are Monte Carlo estimates, small gaps can be sampling noise. We therefore introduce a tolerance $\alpha\in[0,1]$ and label \textsc{pivot} iff
% \[
% \widehat U_1\;<\;\alpha\,\widehat U_0,
% \]
% and \textsc{non-pivot} otherwise. Intuitively, $\alpha$ trades recall vs.\ precision: higher $\alpha$ demands a clearer relative drop to call \textsc{pivot}.

\paragraph{Rollout estimate with tolerance.}
We let $x_t=(\mathbf{x},\mathbf{y}_{<t})$ and candidate $\tilde y_t$. We aim to compute
\[
U_{\text{base}} = U\!\big(p_{\text{target}},\, x_t\big),\qquad
U_{\text{cand}} = U\!\big(p_{\text{target}},\, (x_t,\tilde y_t)\big).
\]
We estimate these by generating $N$ independent rollouts and taking the mean of the binary utility $u$ to obtain $\widehat U_{\text{base}}$ and $\widehat U_{\text{cand}}$. Because both are Monte Carlo estimates, small gaps can be sampling noise. We therefore introduce a tolerance $\alpha\in[0,1]$ and label \textsc{pivot} iff
\[
\widehat U_{\text{cand}} \;<\; \alpha\,\widehat U_{\text{base}},
\]
and \textsc{non-pivot} otherwise. Larger $\alpha$ flags smaller relative drops (higher recall, lower precision).

\paragraph{LLM-as-judge sanity check (patching false negatives).}
Binary utility can miss harmful tokens that “look fine” on average because later steps self-correct or exploit brittle shortcuts (e.g., reward hacking). For candidates initially labeled \textsc{non-pivot}, we inspect the rollouts used to estimate $\widehat U_{\text{cand}}$, i.e., $\mathbf{y}^{(i)}_{> t}\!\sim\! p_{\text{target}}(\cdot\mid x_t,\tilde y_t)$, and collect those with $u=1$ (reached correct answer). We then select the median-length rollout as a representative and prompt an LLM-as-judge to assess reasoning soundness (flagging leaps or contradictions). If the judge deems the reasoning unsound, we flip the label to \textsc{pivot}. This check only flips to \textsc{pivot}, and therefore it cannot introduce unsafe accepts.

\paragraph{Training (what features and objective).}
From each labeled instance we extract features available at SD verification time on the \emph{target} side: (i) the layer-$\ell$ hidden state at position $t$; (ii) the target-model probability of the candidate token; and (iii) the entropy of the target distribution. We then train a small MLP $f_{\text{pivot}}$ to predict \textsc{pivot} vs.\ \textsc{non-pivot}. 

\paragraph{Inference (how it plugs into SD).}
At inference, PAD runs standard SD. Whenever SD would reject at position $t$, we query $f_{\text{pivot}}$ on the target-side features. If $f_{\text{pivot}}$ predicts \textsc{non-pivot} (score $<\sigma$), we override the rejection and accept the draft token. Otherwise, we fall back to the usual SD replacement.
As an extra guardrail, we reject any token whose target-model probability is below $10^{-4}$, regardless of the classifier score. The incremental cost is negligible: a single MLP forward pass on a fixed-size feature vector, which is tiny relative to target/draft transformer steps and easily batched across positions.

% \paragraph{Practical notes.}
% Choice of $N$ and $\alpha$ sets the labeler’s bias–variance point; in practice we tune $(N,\alpha,\sigma)$ jointly to achieve target recall under a fixed labeling budget. The judge is invoked only for provisionally \textsc{non-pivot} candidates with $u\!=\!1$ evidence, keeping overhead low while correcting systematic blind spots of $u$.

See Appendix~\ref{app:method_details} for dataset generation, feature extraction, the MLP architecture, and threshold selection details.

\section{Experiments}

\textbf{Datasets:}
We evaluate our approach on three tasks spanning different domains. For mathematical reasoning, we use GSM8K~\cite{cobbe2021gsm8k}, a dataset of grade-school-level numerical word problems, and AIME24~\cite{huggingfaceh4_aime24}, which consists of problems from the 2024 American Invitational Mathematics Examination. For code generation, we use MBPP (Mostly Basic Python Problems)~\cite{austin2021program}, a collection of crowd-sourced Python programming tasks. For GSM8K and MBPP, we randomly sample 200 test prompts.

\textbf{Generation setting.}
We implement our method on top of the high-performance \texttt{gpt-fast} library~\cite{gptfast}. 
Experiments use the Qwen3 family~\cite{yang2025qwen3}: \texttt{Qwen3-8B} as the target and \texttt{Qwen3-0.6B} as the draft. 
We enable thinking and use decoding parameters \(\text{temperature}=0.6\), \(\text{top\_p}=0.95\), \(\text{top\_k}=20\). 
The maximum context length is 32{,}000 tokens and the speculative length is \(\gamma=10\). 
Experiments are executed on 8\(\times\)A100 GPUs.

\textbf{Metrics:}
For all tasks, we report accuracy (pass@1). For speculative decoding–based methods (PAD and SD), we additionally report the draft acceptance ratio $\eta$ and speedup over target model. Metrics are averaged over 8 completions.

% \textbf{Results}
% \input{tables/evaluation_results}
% Table~\ref{tab:evaluation_results} presents the evaluation results. Note that SD preserves the output distirbution of the target model, hence we report the same accuracy for both. By adjusting the threshold $\sigma$, PAD provides a controllable trade-off between accuracy and efficiency. On the simpler tasks (GSM8K and MBPP), using larger $\sigma$ values achieves greater speedups (up to $2.46\times$ versus $1.57\times$ for SD) while maintaining competitive accuracy. For the more challenging AIME24 task, maintaining high accuracy requires smaller $\sigma$ values, which reduces the achievable speedup relative to SD ($1.95\times$ versus $1.69\times$).

\textbf{Results}
\begin{table}[t]
\centering
\small
\setlength{\tabcolsep}{4pt}
\renewcommand{\arraystretch}{0.95}
\begin{tabular}{l
                c c c  % GSM8k: Acc, eta, Spd
                c c c  % AIME24
                c c c} % MBPP
\toprule
& \multicolumn{3}{c}{\textbf{GSM8k}} & \multicolumn{3}{c}{\textbf{AIME24}} & \multicolumn{3}{c}{\textbf{MBPP}} \\
\cmidrule(lr){2-4}\cmidrule(lr){5-7}\cmidrule(lr){8-10}
\textbf{Setting} & Acc. & $\eta$ (\%) & Spd. & Acc. & $\eta$ (\%) & Spd. & Acc. & $\eta$ (\%) & Spd. \\
\midrule
Target              & $94_{\pm 0.6}$ & —                 & 1.00 & $73_{\pm 4.5}$ & —                 & 1.00 & $70_{\pm 1.9}$ & —                 & 1.00 \\
SD                  &       $94_{\pm 0.6}$        & $45.3_{\pm 0.2}$  & 1.57 & $73_{\pm 4.5}$              & $47.2_{\pm 0.7}$  & 1.69 & $70_{\pm 1.9}$              & $41.8_{\pm 0.4}$  & 1.46 \\
PAD ($\sigma{=}0.7$)& $93_{\pm 1.2}$ & $77.2_{\pm 0.8}$  & 2.46 & $57_{\pm 7.8}$ & $78.8_{\pm 0.5}$  & 2.51 & $64.7_{\pm 1.7}$ & $69.1_{\pm 0.7}$ & 2.25 \\
PAD ($\sigma{=}0.5$)& $93.4_{\pm 0.9}$ & $70.8_{\pm 0.9}$ & 2.33 & $61.6_{\pm 5.3}$ & $71.6_{\pm 0.6}$  & 2.33 & $68.6_{\pm 2.3}$ & $61.7_{\pm 0.2}$ & 2.00 \\
PAD ($\sigma{=}0.3$)& $93.7_{\pm 1.1}$ & $58.2_{\pm 0.2}$ & 1.95 & $69.6_{\pm 4.2}$ & $58.3_{\pm 0.6}$  & 1.95 & $68.3_{\pm 4.8}$ & $50.2_{\pm 0.3}$ & 1.71 \\
Draft               & $74.2_{\pm 1.5}$ & —                & 3.94 & $12.5_{\pm 3.4}$ & —                & 3.94 & $51.1_{\pm 1.3}$ & —                & 3.94 \\
\bottomrule
\end{tabular}
\caption{Evaluation results. Acc.: Accuracy; Spd.: speedup vs.\ target-only; SD: Speculative Decoding; $\eta$: draft acceptance ratio.}
\label{tab:evaluation_results}
\end{table}

Table~\ref{tab:evaluation_results} summarizes accuracy and speedup relative to the target-only baseline. Because SD preserves the target model’s output distribution, its accuracy matches the target’s, so we report the same number for both. PAD introduces a tunable threshold $\sigma$ on the pivot classifier: larger $\sigma$ accepts more draft tokens (higher speedup, potentially lower accuracy), while smaller $\sigma$ is more conservative. On GSM8K and MBPP, PAD shifts the speed–accuracy curve outward: with larger $\sigma$, it attains substantially higher speedups (up to $2.46\times$ versus $1.57\times$ for SD) while retaining competitive accuracy. For the harder AIME24 benchmark, maintaining high accuracy requires smaller $\sigma$, which reduces PAD’s speedup relative to SD ($1.95\times$ versus $1.69\times$). Across settings, the classifier overhead is negligible and gains come from accepting more draft tokens and thus invoking the target model less often.

\section{Related Work}

EAGLE \cite{li2024eagle} and Medusa \cite{cai2024medusa} introduce alternative drafting strategies by leveraging additional decoding heads and specialized draft models, while incorporating verification in parallel with tree attention. Another line of work focuses on designing faster draft models through heuristics \cite{chen2024cascade, he-etal-2024-rest, zhao-etal-2024-ouroboros} or adaptive draft lengths \cite{liu2025pearl}. Others aim to improve draft quality to increase acceptance rates, for example by aligning the draft distribution with the target \cite{zhou2024distillspec} or exploiting features of the target model \cite{du2024glide}. These methods are orthogonal to our work, which instead emphasizes the verification phase.
Closer to our approach, some methods relax strict distribution matching to improve acceptance. \citet{kim2023speculative} leverages token-level uncertainty, while \cite{bachmann2025judge} trains a classifier to accept or reject tokens. However, both rely on heuristic criteria rather than a principled utility-based formulation. Moreover, \citet{bachmann2025judge} depends on hand-crafted datasets. By contrast, we reformulate decoding as a utility-based objective, yielding a fully self-supervised method grounded in task performance rather than manually designed criteria. More recently, other approaches \cite{liao2025reward, fu2025scaling, pan2025specreason} have proposed combining draft and target model outputs at the step level. However, these methods depend on an auxiliary reward model or process for scoring, which introduces additional overhead.

\section{Conclusion}

In this work, we proposed a novel reformulation of Speculative Decoding that focuses on preserving the utility of the target model rather than strictly matching its sampling distribution. Specifically, we introduced Pivot-Aware Speculative Decoding, a decoding strategy that rejects only the pivotal tokens, referred to as \textit{pivot tokens}, which are likely to lead to a utility drop in the final output. To enable this, we trained a lightweight classifier to detect pivot tokens. Our method achieves a substantial speedup, up to $2.51 \times$, while maintaining the performance and utility of the target model.

\section{Acknowledgments}

This work was supported by Samsung Electronics (Samsung Semiconductor, USA). The opinions, findings, and conclusions or recommendations expressed are those of the author(s) and do not necessarily reflect the views of the sponsor.

% \section{Limitations}

\bibliography{custom}

\appendix

\section{Technical Appendices and Supplementary Material}
Technical appendices with additional results, figures, graphs and proofs may be submitted with the paper submission before the full submission deadline (see above), or as a separate PDF in the ZIP file below before the supplementary material deadline. There is no page limit for the technical appendices.

\subsection{Proof of Lemma~\ref{lem:recall} }\label{sec_proof}
\begin{proof}[\textbf{Proof of Lemma~\ref{lem:recall}}]

Let $f_{\text{pivot}}(x_{ot}, \mathbf{x}_{ot<}, \mathbf{x}_c)$ be a binary classifier with 100\% recall, such that it correctly identifies all pivot tokens for any prefix $\mathbf{x}_{ot<}$ and context $\mathbf{x}_c$. We aim to show that, for any $\mathbf{x}_c$, the utility of the sampled output with any random seed $s$ from the target model, $u(p^{s}_{\text{target}}(\cdot \mid \mathbf{x}_c), \mathbf{x}_c)$, is equal to the utility of the sampled output with any random seed from the utility-aware decoding strategy, $u(p^{s}_{ad}(\cdot \mid \mathbf{x}_c), \mathbf{x}_c)$. This would naturally imply that the objective in Equation~\ref{objective} is satisfied with $\epsilon = 0$.

We prove this by induction.

Define a completion function $C_s(\mathbf{x}_{\text{gen}}, \mathbf{x}_c)$ with seed $s$ which, given a partially generated token sequence $\mathbf{x}_{\text{gen}}$ and context $\mathbf{x}_c$, returns the full sequence by completing $\mathbf{x}_{\text{gen}}$ using $p_{\text{target}}$, unless $\mathbf{x}_{\text{gen}}$ is already a completed sequence, in which case it returns $\mathbf{x}_{\text{gen}}$ directly.

Our goal is to show that for any generation prefix $\mathbf{x}_{ot}$ produced by $p_{ad}$,
\[
u(C_s(\mathbf{x}_{ot}, \mathbf{x}_c), \mathbf{x}_c) \geq u(p^{s}_{\text{target}}(\cdot \mid \mathbf{x}_c), \mathbf{x}_c), \forall t,s.
\]
If this holds at every token generation step, it follows that the final utility of utility-aware decoding is lower bounded by that of the target model:
\[
u(p^{s}_{ad}(\cdot \mid \mathbf{x}_c), \mathbf{x}_c) \geq u(p^{s}_{\text{target}}(\cdot \mid \mathbf{x}_c), \mathbf{x}_c),
\]
since the full generation of $p_{ad}$ corresponds to the final step.

\textbf{Base Case:}  
When no tokens have yet been generated, i.e., $\mathbf{x}_{ot} = ()$, we have:
\[
u(C_s((), \mathbf{x}_c), \mathbf{x}_c) \geq  u(p^{s}_{\text{target}}(\cdot \mid \mathbf{x}_c), \mathbf{x}_c)
\]
by the definition of the completion function $C$ which implies actually the equality.

\textbf{Inductive Step:}  
Assume that for some prefix $\mathbf{x}_{ot}$ up to step $t$, the condition holds:
\[
u(C_s(\mathbf{x}_{ot}, \mathbf{x}_c), \mathbf{x}_c) \geq u(p^{s}_{\text{target}}(\cdot \mid \mathbf{x}_c), \mathbf{x}_c).
\]
Now consider the generation of the next token $x_{ot+1}$. If $x_{ot+1}$ is accepted by $p_{ad}$, it must not be an pivot token. By the definition of pivot tokens, adding $x_{ot+1}$ does not decrease the final utility of $u(C(\mathbf{x}_{ot}, \mathbf{x}_c), \mathbf{x}_c )$, which is greater-equal to $u(p^{s}_{\text{target}, \mathbf{x}_c}(\cdot \mid \mathbf{x}_c), \mathbf{x}_c)$ by the inductive assumption. If the token is rejected, it is replaced by the token sampled from $p^s_{\text{target}}$, which does not decrease the utility due to the inductive assumption and the definition of the completion function $C_s$. Specifically, we have:
\[
u(p^{s}_{\text{target}}(\cdot \mid \mathbf{x}_c), \mathbf{x}_c) \leq u(C_s(\mathbf{x}_{ot}, \mathbf{x}_c), \mathbf{x}_c) \leq  u(C_s((x_{ot+1}, \mathbf{x}_{ot}), \mathbf{x}_c), \mathbf{x}_c),
\]
where $x_{ot+1}$ is the next token selected by $p_{\text{target}}$ with seed $s$.
 Therefore:
\[
u(C_s(\mathbf{x}_{ot+1}, \mathbf{x}_c)) \geq u(p^{s}_{\text{target}}(\cdot \mid \mathbf{x}_c)).
\]

As this proof is for any sampling with any random seed $s$ and by induction, utility-aware decoding produces an output with the same utility as the target model for all $\mathbf{x}_c$, satisfying Equation~\ref{objective} which is an expectation over all possible seeds $s$,    with $\epsilon = 0$.

\end{proof}
%%%%%%%%%%%%%%%%%%%%%%%%%%%%%%%%%%%%%%%%%%%%%%%%%%%%%%%%%%%%

% \subsection{Interpretability Prompt}\label{interpretability_prompt}

% \subsection{Dataset Generation}

\section{Dataset Generation and Pivot Classifier} \label{app:method_details}
\begin{figure}[!t]
  \centering
  % replace the path below with your project path if different
  \includegraphics[width=0.5\columnwidth]{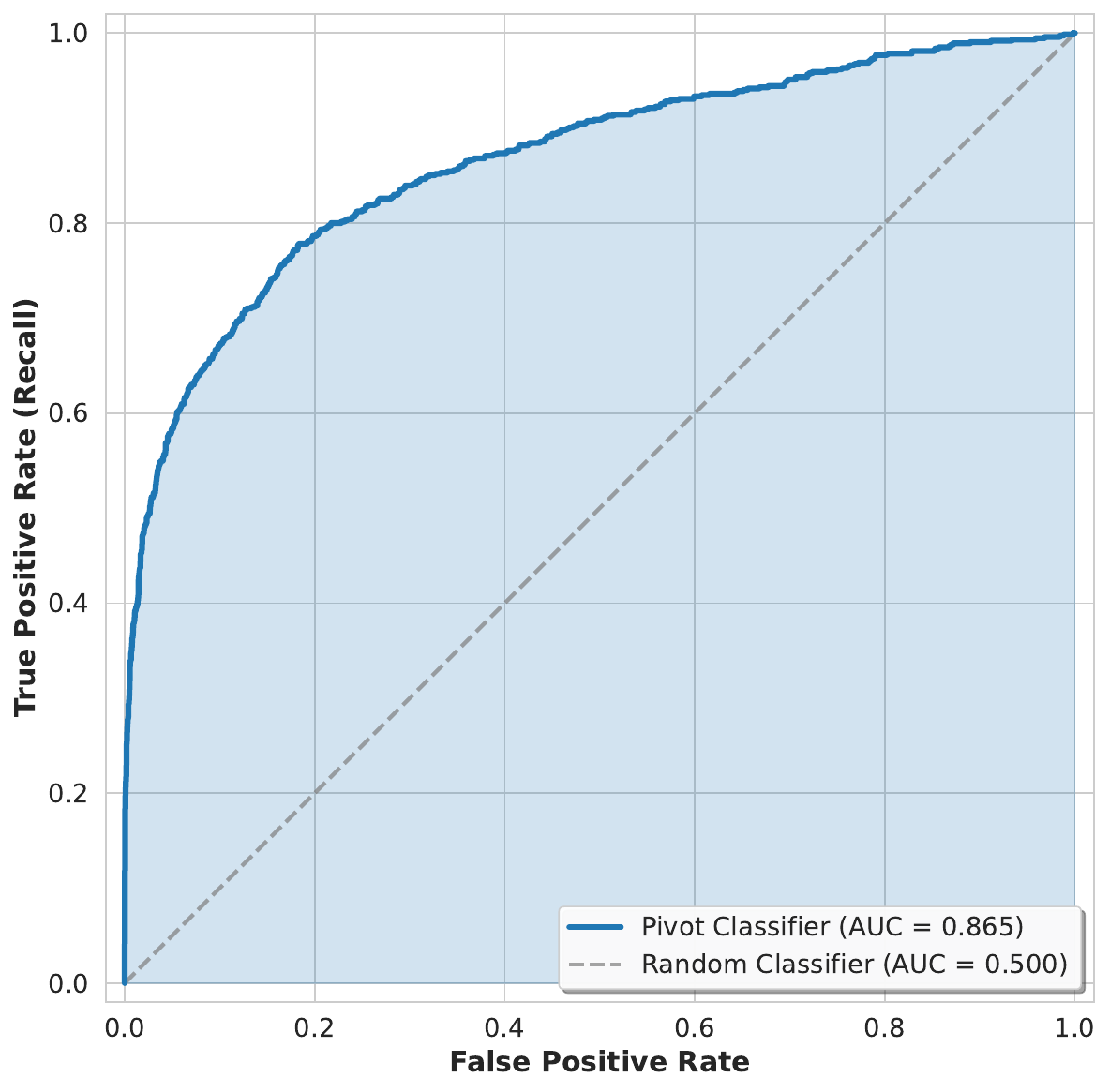}
  \caption{ROC curve for the pivot classifier on the held-out test set.}
  \label{fig:pivot-roc}
\end{figure}

\paragraph{Dataset generation.}
Algorithm \ref{alg:dataset_gen} details the data collection procedure described in Section \ref{sec:classifier}. We use the GSM8K training set~\cite{cobbe2021gsm8k} and generate $32619$ labeled samples, with $3657$ tokens labeled as \emph{pivot} and $28962$ as \emph{non-pivot}. The target model is \texttt{Qwen3-8b} and the draft model is \texttt{Qwen3-0.6b}. We disable “thinking” to keep rollout costs manageable. Also, $\alpha$ is set to $0.8$. For the soundness check, we employ \texttt{Gemini-Flash-Light} as the LLM-as-judge with medium reasoning effort\footnote{\href{https://ai.google.dev/gemini-api}{Gemini
 API}: \texttt{Gemini-Flash-Light}}. The prompt used for the soundness check is provided below (Prompt B.1).

\paragraph{Pivot classifier.}
We predict whether a draft token is a pivot using an MLP over the target hidden state at layer $l$ ($\mathbf{h}_l$), token entropy $H$, and the target-model probability of the candidate token $p$. Let $\mathbf{s}=[H,\,p]$ and define the concatenation operator $\oplus$ (i.e., $a \oplus b$ stacks vectors $a$ and $b$). With ReLU $\phi(\cdot)$, and writing $\mathrm{Linear}(x)=Wx+b$, the model is:
\[
\begin{aligned}
\mathbf{u} &= \phi\!\big(\mathrm{Linear}_{h}(\mathbf{h}_l)\big),\\
\mathbf{v} &= \phi\!\big(\mathrm{Linear}_{s}(\mathbf{s})\big),\\
\mathbf{c} &= \mathbf{u} \oplus \mathbf{v},\\
\boldsymbol{z} &= \mathrm{Linear}_{\text{out}}\!\Big(\phi\big(\mathrm{Linear}_{\text{fuse}}(\mathbf{c})\big)\Big),\quad \boldsymbol{z}\in\mathbb{R}^{2},\\
\hat{\boldsymbol{y}} &= \mathrm{softmax}(\boldsymbol{z}),
\end{aligned}
\]
where $\mathrm{Linear}_{\text{fuse}}$ is a standard linear layer applied to the concatenated vector $\mathbf{c}$, and $\mathrm{Linear}_{\text{out}}$ maps the fused features to two logits (pivot / non-pivot). To address class imbalance, we train with weighted cross-entropy, split the data 80/20 into train/test, and select the checkpoint with the lowest validation loss. The ROC curve, obtained by sweeping the decision threshold $\sigma$, yields $\text{AUC}=0.865$, well above the $0.5$ random baseline, indicating robust separation of pivot and non-pivot tokens across thresholds (Fig.~\ref{fig:pivot-roc}). At deployment, $\sigma$ can be tuned to trade off acceptance rate against target-model fallbacks to meet a given latency/utility budget.

\begin{algorithm*}[t]
\caption{Pivot Classifier Data Collection}
\label{alg:dataset_gen}
\KwIn{$\mathcal{D}$ (prompts $x$), target $p_t$, draft $p_d$, $\mathrm{SD\_ACCEPT}$, $\mathrm{SD\_RESAMPLE}$, binary utility $u$, rollouts $N$, tolerance $\alpha$, judge $\mathrm{JudgeIsSound}$, steps $T$}
\KwOut{$\mathcal{S}=\{(\text{text},\ell)\}$ with $\ell\in\{\texttt{pivot},\texttt{non-pivot}\}$ \quad (features extracted later)}

\SetKwFunction{Roll}{Rollouts}
\SetKwFunction{Sel}{SelectRep}
\SetKwProg{Fn}{func}{:}{}

% ----- Utility definition (true vs. MC estimate) -----
% U(x,y_{<t}) = E_{y \sim p_t(\cdot|x,y_{<t})}[u(y,x)]
% \hat U(x,y_{<t}) = (1/N) \sum_{n=1}^N u(y^{(n)},x),\; y^{(n)} \sim p_t(\cdot|x,y_{<t})

% Rollouts: return utilities and lengths for later selection
\Fn{\Roll{$x,y_{<t}$}}{
  sample $N$ completions $y^{(n)}\sim p_t(\cdot\mid x,y_{<t})$; \\
  \Return $\{(y^{(n)},\,u(y^{(n)},x),\,|y^{(n)}|)\}_{n=1}^N$
}

% Representative chooser: median length among positives
\Fn{\Sel{$\mathcal{P}$}}{
  \tcp{$\mathcal{P}=\{(y, u{=}1, \ell)\}$}
  sort $\mathcal{P}$ by sequence length $\ell$; \Return middle element (median length)
}

$\mathcal{S}\leftarrow\emptyset$\;
\For{$x\sim\mathcal{D}$}{
  initialize prefix $y_{<1}\leftarrow\emptyset$\;
  \For{$t=1$ \KwTo $T$}{
    \tcp{1) Draft proposes next token}
    sample $\tilde y_t\sim p_d(\cdot\mid x,y_{<t})$\;

    \tcp{2) If SD would accept, append and move on (not a candidate)}
    \If{$\mathrm{SD\_ACCEPT}(x,y_{<t},\tilde y_t,p_t)$}{
        $y_{<t+1}\leftarrow (y_{<t},\tilde y_t)$; \textbf{continue}
    }

    \tcp{3) Otherwise, evaluate MC utilities for base vs. candidate}
    $\mathcal{B}\leftarrow$\Roll{$x,\,y_{<t}$}; \quad
    $\hat U_{\mathrm{base}}\leftarrow \frac{1}{|\mathcal{B}|}\sum_{(y,u,*)\in\mathcal{B}} u$\;
    $\mathcal{C}\leftarrow$\Roll{$x,\,(y_{<t},\tilde y_t)$}; \quad
    $\hat U_{\mathrm{cand}}\leftarrow \frac{1}{|\mathcal{C}|}\sum_{(y,u,*)\in\mathcal{C}} u$\;

    \tcp{4) Label as pivot / non-pivot (with judge check for false wins)}
    \eIf{$\hat U_{\mathrm{cand}}<\alpha\,\hat U_{\mathrm{base}}$}{
      $\ell\leftarrow\texttt{pivot}$
    }{
      $\ell\leftarrow\texttt{non-pivot}$\;
      $\mathcal{P}\leftarrow\{(y,u,\ell): (y,u,\ell)\in\mathcal{C},\ u=1\}$\;
      \If{$\mathcal{P}\neq\emptyset$}{
        $y^{+}\leftarrow$\Sel{$\mathcal{P}$}\;
        \If{$\neg\,\mathrm{JudgeIsSound}\big(\mathrm{concat}(x,\,y^{+})\big)$}{
          $\ell\leftarrow\texttt{pivot}$
        }
      }
    }

    \tcp{5) Persist labeled sample for training (only when SD rejected)}
    $\text{text}\leftarrow \mathrm{concat}\big(x,\,(y_{<t},\tilde y_t)\big)$\;
    append $(\text{text},\ell)$ to $\mathcal{S}$\;

    \tcp{6) Update prefix for next step per spec}
    \eIf{$\ell=\texttt{non-pivot}$}{
        $y_{<t+1}\leftarrow (y_{<t},\tilde y_t)$ \tcp*{accepted for next step}
    }{
        \tcp{pivotal: fall back to SD rejection path (target-side resampling)}
        $y_t \leftarrow \mathrm{SD\_RESAMPLE}(x,y_{<t},p_t)$\;
        $y_{<t+1}\leftarrow (y_{<t},y_t)$
    }
  }
}
\Return $\mathcal{S}$ \tcp*{(run offline feature extraction next)}
\end{algorithm*}

\begin{fullwidthprompt}
You are an expert evaluator specializing in identifying flawed reasoning in problem-solving narratives. Your task is to analyze a given response and determine if it contains logical errors, even if the final answer is correct.

Read the entire 'Response to Analyze' from start to finish. A response is considered to have flawed reasoning if you identify any of the following patterns:

- **Incorrect Intermediate Steps:** It contains calculation errors or incorrect logical steps that are later implicitly corrected or ignored to reach the correct final answer.
- **Logical Fallacies:** The reasoning contains leaps of logic or doesn't coherently lead to the conclusion, but the result coincidentally matches the correct answer.
- **Self-Correction:** The response states an incorrect piece of information or result and then corrects itself. Sometimes (not always) there are explicit phrases like "Wait, that's not right," "But the question asks for...," or "Let's try again" before the correct path is found.
- **Unnecessary or Redundant Steps:** It includes steps or calculations that are irrelevant and do not contribute to the final answer, indicating a confused or inefficient thought process.
- **Other Reasoning Flaws:** Any other type of faulty logic that raises questions about the soundness of the process.

---

### **Examples**

**EXAMPLE 1: FLAWED REASONING (Self-Correction)**

* **Question:** A bat and a ball cost \$1.10 in total. The bat costs \$1.00 more than the ball. How much does the ball cost?
* **Response to Analyze:** The bat costs \$1.00, so the ball must cost \$0.10. Wait, that's not right because the difference is only \$0.90. Let's try again. If the ball is B, the bat is B + 1. So B + (B + 1) = 1.10, which means 2B = 0.10. The ball costs \$0.05.
* **Your JSON Output:**
    ```json
    {{
      "analysis": "The response initially presents an incorrect answer (\$0.10) but then immediately identifies the error and uses a correct algebraic method to perform a self-correction, arriving at the right answer.",
      "decision": true
    }}
    ```

**EXAMPLE 2: SOUND REASONING (No Flaws)**

* **Question:** A bakery has 5 boxes of donuts, with 12 donuts in each box. They sell 3 boxes. How many donuts are left?
* **Response to Analyze:** First, determine the number of boxes remaining, which is 5 - 3 = 2 boxes. Then, calculate the total donuts in the remaining boxes: 2 boxes * 12 donuts/box = 24 donuts.
* **Your JSON Output:**
    ```json
    {{
      "analysis": "The reasoning is sound, direct, and efficient. It correctly calculates the remaining boxes before finding the total number of donuts left. There are no logical flaws.",
      "decision": false
    }}
    ```

**EXAMPLE 3: FLAWED REASONING (Unnecessary Steps)**

* **Question:** A recipe requires 2 cups of flour to make 12 cookies. You want to make 36 cookies. How much flour do you need?
* **Response to Analyze:** To make 36 cookies, which is 3 times 12, you'll need 3 times the flour. So, 2 cups * 3 = 6 cups. The oven should be preheated to 350F. The total flour needed is 6 cups.
* **Your JSON Output:**
    ```json
    {{
      "analysis": "The core calculation is correct, but the response introduces an unnecessary and irrelevant piece of information ('The oven should be preheated to 350F') that does not contribute to solving the problem.",
      "decision": true
    }}
    ```

---

### **Input Data for Analysis**

**1. Question:**
{question}

**2. Golden Answer (for your reference):**
{ground_truth}

**3. Response to Analyze:**
{response}

---

### **Your Task**

After carefully reviewing the 'Response to Analyze', provide your evaluation in a strict JSON format, following the structure shown in the examples above. The JSON object must contain exactly two keys: `"analysis"` and `"decision"`.

**Your JSON Output:**
\end{fullwidthprompt}

\end{document}